\documentclass[letterpaper, 10pt, conference]{ieeeconf}  

\IEEEoverridecommandlockouts                              

\usepackage{graphics} 
\usepackage{epsfig} 
\usepackage{amsmath} 
\usepackage{cite}
\usepackage{graphics} 
\usepackage{epsfig} 
\usepackage{mathptmx} 
\usepackage{times} 
\usepackage{amsmath} 
\usepackage{amssymb}  
\usepackage{subfigure}
\usepackage{algorithm} 
\usepackage{algpseudocode}
\usepackage{multirow} 
\usepackage{rotating}
\usepackage{tablefootnote}
\usepackage{threeparttable}
\usepackage{balance}
\usepackage{booktabs}
\usepackage{bookmark}
\usepackage{makecell}
\usepackage{array}
\usepackage{caption}
\usepackage{hyperref}
\hypersetup{hidelinks,
	colorlinks=true,
	allcolors=black,
	pdfstartview=Fit,
	breaklinks=true}

\title{\LARGE \bf
Foresee and Act Ahead: Task Prediction and Pre-Scheduling Enabled Efficient Robotic Warehousing
}

\author{Bo Cao, Zhe Liu, Xingyao Han, Shunbo Zhou, Heng Zhang, Hesheng Wang
\thanks{B. Cao and H. Wang are with Department of Automation, Shanghai Jiao Tong University, China. Z. Liu and X. Han are with the MoE Key Lab of Artificial Intelligence, AI Institute, Shanghai Jiao Tong University, China. S. Zhou
and H. Zhang are with the Edge Cloud Innovation Lab, Huawei Cloud
Computing Technologies Co., Ltd. This work was supported by Huawei Cloud Computing Technologies Co., Ltd. under the Project 9450529.   
}
}

\begin{document}

\maketitle
\thispagestyle{empty}
\pagestyle{empty}

\begin{abstract}
In warehousing systems, to enhance logistical efficiency amid surging demand volumes, much focus is placed on how to reasonably allocate tasks to robots. However, the robots labor is still inevitably wasted to some extent. In response to this, we propose a pre-scheduling enhanced warehousing framework that predicts task flow and acts in advance. It consists of task flow prediction and hybrid tasks allocation. For task prediction, we notice that it is possible to provide a spatio-temporal representation of task flow, so we introduce a periodicity-decoupled mechanism tailored for the generation patterns of aggregated orders, and then further extract spatial features of task distribution with novel combination of graph structures. 
In hybrid tasks allocation, we consider the known tasks and predicted future tasks simultaneously and optimize the allocation dynamically. In addition, we consider factors such as predicted task uncertainty and sector-level efficiency evaluation in warehousing to realize more balanced and rational allocations.
We validate our task prediction model across actual datasets derived from real factories, achieving SOTA performance. Furthermore, we implement our compelte scheduling system in a real-world robotic warehouse for months of lifelong validation, demonstrating large improvements in key metrics of warehousing, such as empty running rate, by more than 50\%.
\end{abstract}

\section{Introduction}
To improve the efficiency of warehousing system,
the problem of Multi-Robot Task Allocation (MRTA) remains a prominent research topic. Numerous task allocation approaches have been proposed and shown to be effective\cite{mrta_review, mrta_review2,mrta_review3}. The allocation methods include market-based, behavior-based, and optimization-based approaches, in which some also take into account the uncertainty when executing to dynamically assign for robots\cite{mrta_app1}. However, these methods are invariably developed based on existing tasks or other information, leading to two main issues: 1) The task execution strategies of multi-robot systems (MRS) largely depends on current demand, which limits the potential operational capacity of the warehousing. 2) In specific situations, such as during off-peak seasons, robotic labor utilization is significantly reduced, resulting in resource waste.

\begin{figure}[!t]
\centering
{\includegraphics[width=1\columnwidth]{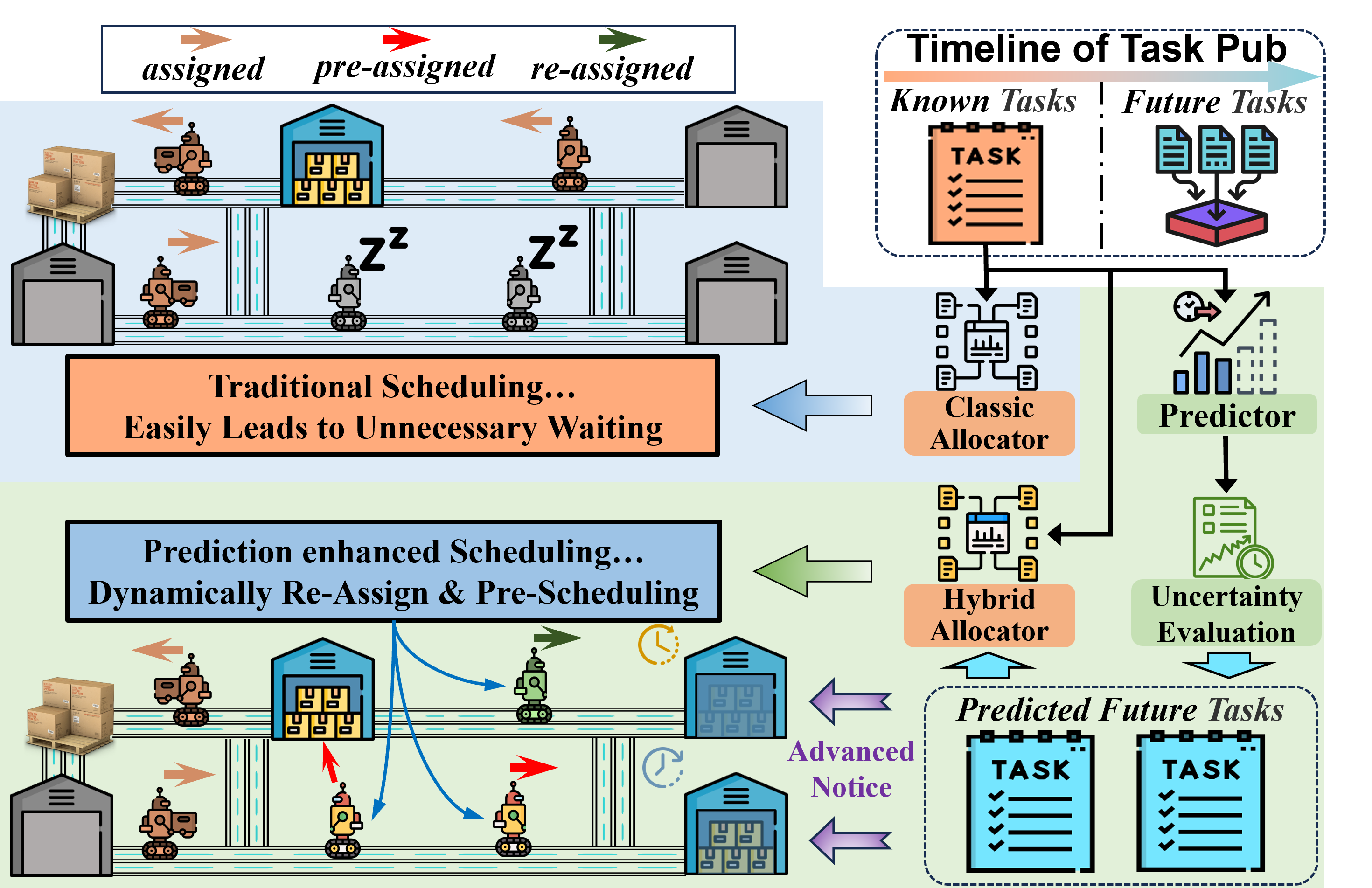}}
\caption{The pre-scheduling enhanced framework for robotic warehousing. It predicts future tasks along with the timeline while assessing the uncertainties, following which, a hybrid allocator is designed to allocate hybrid tasks. This allows otherwise idle robots to move ahead and be dynamically re-assigned them in each round of allocation, further exploring the potential of robotic labor in multi-robot warehousing system.} 
\label{fig: intro_pic}
\vspace{-0.5cm}
\end{figure}

Nowadays, with the development of deep learning, particularly in spatio-temporal prediction, data-driven methods are applied to MRS to uncover hidden information\cite{our_pred, han_1}, thereby improving collaboration between robots. However, to date, no work has leveraged the estimation and prediction of task distribution to enable more effective management and scheduling of warehouse logistics based on both predicted and existing tasks. To take advantage of this, our primary idea is to predict task flow by capturing the spatio-temporal dependencies of tasks in warehousing. Then these predicted tasks are treated as special candidate tasks in the task pool and are allocated to robots in a more rational manner. A conceptual diagram of the framework is shown as Fig.~\ref{fig: intro_pic}. 

In fields such as traffic flow prediction and ride-hailing demand forecasting, there have been brilliant applications of spatio-temporal prediction approaches\cite{st_app_1, taxi_1}. In warehousing, however, task flow is often sparsely and unevenly distributed, and the adjacency relations among areas where tasks emerge are complex due to diverse delivery demand. These issues make it hard to characterize the distribution of task flow and to predict it with a satisfied accuracy. Therefore, the robotic warehousing domain urgently requires a scenario-adaptive approach to predict task flow. Additionally, how to rationally allocate the hybrid tasks still remains a problem.

Our contributions are as follows: 
 1) We introduce, for the first time, a pre-scheduling enhanced framework for multi-robot warehousing system. It enables rolling prediction of task flow distributions and optimizes task allocation patterns, maximizing the efficiency and utilization rate of robotic labor. 
 2) We propose a novel network TDTGCN, which extracts the spatio-temporal dependencies of task distribution through the fusion of Classic Graph, HyperGraph, and Heterogeneous Graph structures, improving predictive accuracy. 
 3) Real-world robotic warehouse implementations are developed with hundreds of robots to validate the feasibility. Months of lifelong validations indicate that we improves relevant metrics significantly by more than 50\%.

\vspace{-6pt}
\section{Related Works}
\subsection{Multi-Robot Task Allocation}
Various MRTA methodologies have been developed for different MRS application scenarios \cite{ha, behavior_based_1, market_based_1}. However, most existing approaches focus on optimally distributing tasks from a predefined pool, often overlooking the impact of the task distribution mechanism itself. This oversight is particularly critical in structured environments like warehouse logistics, where a surge of orders can lead to intractable situations. Our work addresses this gap by integrating prior information about task distribution into the allocation mechanism, thereby enhancing the coordinated management of robots in MRS.

\subsection{Spatio-Temporal Prediction}
Spatio-temporal prediction has been widely applied, such as urban traffic flow prediction\cite{traffic_flow} and ride-hailing demand prediction\cite{ridehailing}. Early predictive models, such as GBRT and other hierarchical inference models\cite{inference_model}, struggle with poor generalizability and high computational demands. With the rise of deep learning, models like RNNs, Transformers, Diffusions, and Graph Neural Networks (GNNs) have shown great promise in capturing spatial dependencies and underlying patterns\cite{st_review1,st_review2}. Notable approaches include ASTGCN\cite{astgcn}, DCRNN\cite{dcrnn} and STRGAT\cite{strgat} which leverage GNNs for spatial representation and information aggregation. Mechanisms like dilated convolution\cite{gwn} and spatio-temporal attention\cite{attentiongcn} further boost the ability to capture latent dependencies and improve generalization.

\subsection{Challenges of Task Flow Prediction in Warehousing}\label{subsec: vs}
In warehousing, the environment can be analogized to a structured urban setting, with delivery tasks representing a form of demand. However, significant gaps remain between those applications and warehousing: 1) \textit{Sparsity of Task Distribution}: Warehousing tasks often exhibit temporal sparsity, with minimal variation in data flow across multiple sampling periods. 2) \textit{Complex Regional Adjacency Relations}: Unlike applications like taxi demand prediction, where specific areas are modeled as graph nodes and dependencies are typically extracted using methods such as GraphSAGE\cite{graphsage} or Graph Attention Networks, warehousing presents more complex spatial relationships. These complexities necessitate the construction of additional relationships, such as the direction of connecting edges (representing cargo delivery routes) and various types of edges (indicating specific delivery requirements). To the best of the authors' knowledge, there is no existing works which address the above issues. 


\section{System Formulation}

\subsection{Task Flow Prediction}

\subsubsection{Representation of Spatial-Temporal Task Flow}
In the warehouse, roadmap $G=(V, E)$ is a topology graph, where $V$ denotes vertices and $E$ represents edges. The spatio-temporal task flow is encoded into a three-dimensional tensor $\textbf{X} \in \mathbb{R}^{\lambda \times T \times F}$. Here, $\lambda$ is the number of sectors $S_i$ in $G$, where $S = \{S_1, S_2, \dots, S_{\lambda}\}$. Each sector corresponding to a subgraph $G_s(V_s, E_s)$ with $V_s \subseteq V$ and $E_s \subseteq E$, where $|V_s| = \lambda, |E_s| = \gamma$. They form a high-level weighted directed graph $G_H = (V_H, E_H, \textbf{A})$. $T$ denotes the number of data acquisition frames, with each frame captured at fixed intervals, and $F$ indicates the number of features in each sector. Each $\textbf{X}_{s,t,f}$ represents the $f$-th feature of the $s$-th sector at the $t$-th frame.

\subsubsection{The Prediction Process}
We regard the Task Flow Prediction (TFP) as an autoregressive process. In this work, the input data consists of a sequence of tasks at $I$ time steps and \textit{$\lambda$} sector locations. We define the adjacency matrix among secotrs as $\textbf{A} \in \mathbb{R}^{\lambda \times \lambda}$, specifically defined as:
\begin{equation}
    \vspace{-2pt}
    \textbf{A}_{ij}=\left\{\begin{array}{ll} \exp(-\frac{d_{ij}^2}{\sigma^2}) ,\text{  }if \text{  }\exp(-\frac{d_{ij}^2}{\sigma^2})\geq\epsilon\text{ and }i\neq j\\0, \quad\quad\quad\text{otherwise.}\end{array}\right.
    \vspace{-2pt}
\end{equation}
where $\sigma$ is a normalization parameter, and $\epsilon$ is a threshold that controls the sparsity of the adjacency matrix. At time step $k$, the task flow data can be represented as $\textbf{X}^k_\lambda=[x_1^k,x_2^k,...,x_\lambda^k]^T \in \mathbb{R}^{\lambda \times F}$. The input containing the $I$-frame data can be represented as $\textbf{X}^{[(k-I+1):k]}_\lambda=[\textbf{X}^{k-I+1}_\lambda,\textbf{X}^{k-I}_\lambda,...,\textbf{X}^{k}_\lambda]$. The goal of TFP is to construct a mapping $g$ that is able to predict the future $O$-frame task flow data based the historical task flow information of $I$-frames and the adjacency relationship among $\lambda$ sectors. This mapping relationship is represented as: $[\textbf{X}_\lambda^{[(k-I+1):k]}, G_H] \xrightarrow{g} \hat{\textbf{X}}_\lambda^{[k:k+O]}$, where $\textbf{X}_\lambda^{[(k-I+1):k]}\in \mathbb{R}^{\lambda \times I \times F}$ and $\hat{\textbf{X}}_\lambda^{[k:k+O]}\in \mathbb{R}^{\lambda \times O \times F}$.

\subsection{Multi-Robot Hybrid Tasks Allocation (MR-HTA)}
In this work, the definition multi-robot hybrid tasks allocation is based on the following settings:
\begin{itemize}
  \item Consider a set of robots $R$, where $R = \{r_1, r_2, \dots, r_n\}$, and each robot $r_i \ (i=1,2,\dots,n)$ is restricted to performing only one task at any given time. The set of tasks is denoted by $T = \{T_1, T_2, \dots, T_m\}$. 
  \item In contrast to conventional MRTA, the task set $T$ comprises $l$ real tasks, denoted by $T^{\text{real}}$, and $m\text{-}l$ predicted tasks, denoted by $T^{\text{pred}}$.
\end{itemize}At each time step $k$ for tasks allocation, each real task $T^{real}_k$ is associated with a specific vertex $v_i^{real} \in V$ in $G$. In contrast, each predicted task $T^{pred}_k$ is associated with a sector $S_j$, represented by its center $v_{S_j}^{pred}$.

The objective of MR-HTA is to design a funtion $Cost = \mathcal{F}(R, T)$  that optimally matches the $n$ robots with the $m$ tasks (distinguishing between real and predicted tasks) while minimizing the $Cost$. It contains not only cost of completing tasks, but also the prediction confidence score $c_k$ and the given sector's current task completion rate $\eta_{S_j}$. The MR-HTA involves associating each robot vertex $v_r$ with either a $v_i^{real} \in V^{real}$ or a $v_{S_j}^{pred} \in V^{pred}$. Every single match $M$ can be expressed as:
\vspace{-0.1cm}
\begin{equation}
\begin{aligned}
M \subseteq & \{(r_i, v_j^{real}) \mid r_i \in R, v_j^{real} \in V^{real}\} \\
& \cup \{(r_i, v_{S_k}^{pred}) \mid r_i \in R, v_{S_k}^{pred} \in V^{pred}\}
\end{aligned}
\end{equation}

\subsection{System Framework}
\begin{figure}[htpb]
    \centering
    \includegraphics[width=0.85\linewidth]{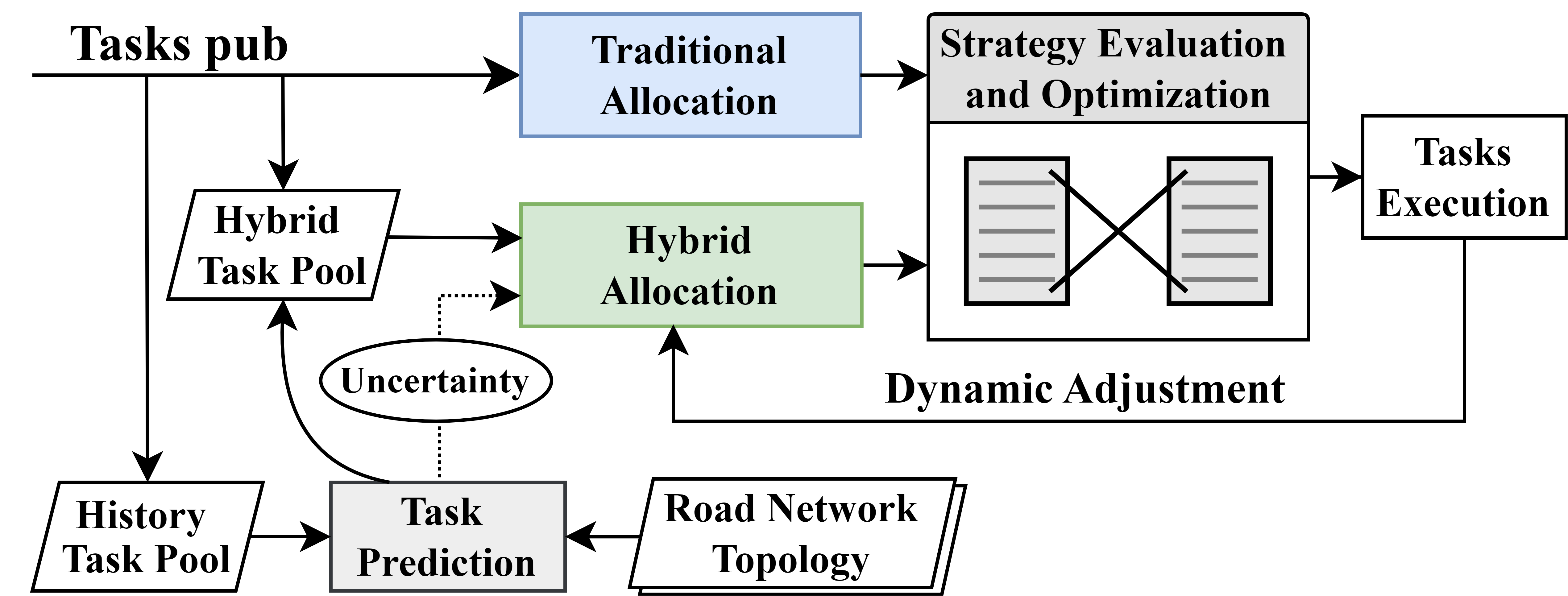}
    \caption{The execution process of our complete scheduling framework.} 
    \label{fig:framework}
\end{figure}

Our framework is depicted in Fig.~\ref{fig:framework}. The system is capable of directly allocating tasks that have been published. Additionally, by leveraging the historical task pool and the topological structure of the high-level $G_H$, it predicts the distribution of future task flow to enable more rational and efficient task allocation. After this, the system evaluates and globally optimizes the task allocation strategy to make a final decision. Moreover, the allocation strategy is dynamically adjusted based on task execution performance to maintain dynamic equilibrium within the warehousing system.

\vspace{-4pt}




\vspace{-0.4cm}
\section{Main Approach}

\subsection{Multi-Robot Task flow Prediction}\label{sec: TFP}
\begin{figure*}[!t]
    \centering
    \includegraphics[width=1.85\columnwidth]{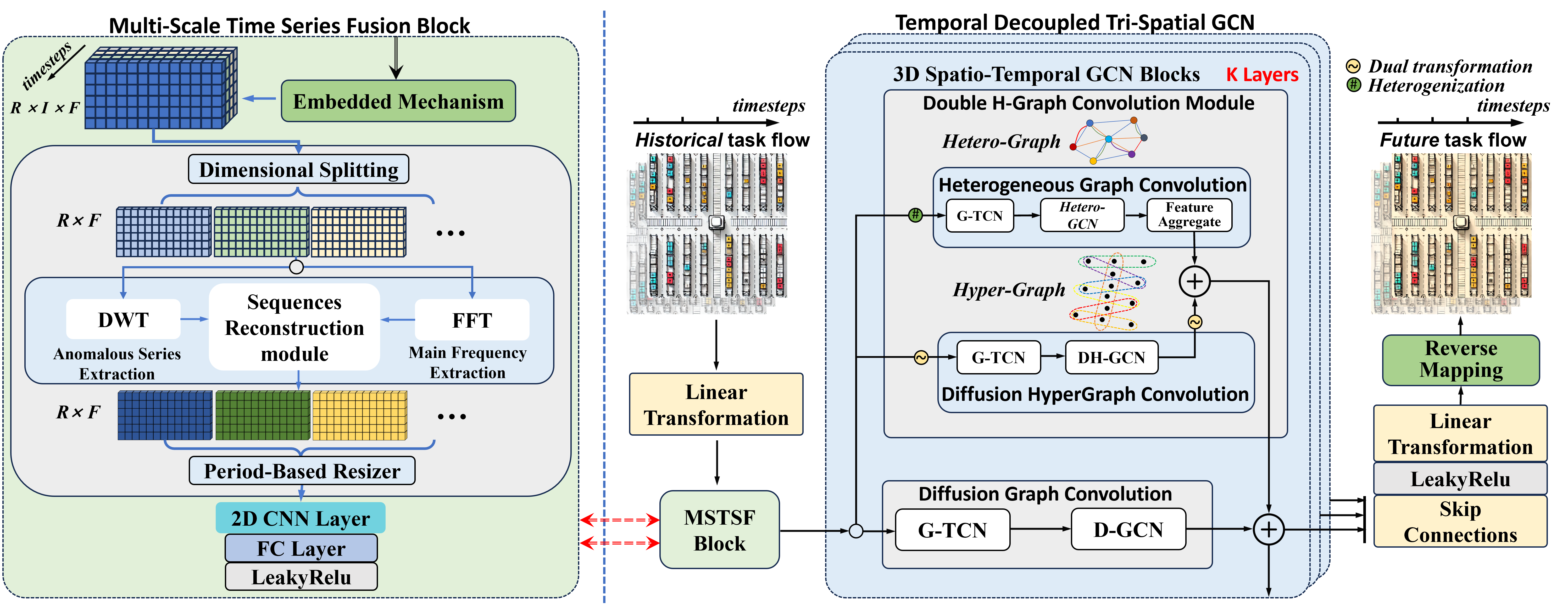}
    \caption{The architecture of the TDTGCN (shown on the right), main blocks connected by residuals. It processes historical task flow data $X_{h} \in \mathbb{R}^{R \times I \times F}$, initially through the MSTSF Block (as depicted on the left). This block embeds sparse data and decouples time-series patterns using multi-scale DWT and FFT. Subsequently, the data is processed by 3D Spatio-Temporal GCN Blocks, extracting spatio-temporal dependencies across 3 spatial dimensions. Finally, the features are integrated and a reverse mapping operation is performed to generate the predicted task flow $X_{p} \in \mathbb{R}^{R \times T' \times F}$.
    }
    \label{fig: tdtgcn}
    \vspace{-0.6cm}
\end{figure*}

As shown in Fig.~\ref{fig: tdtgcn}, we propose Temporal Decoupled Tri-Spatial Graph Convolutional Network (TDTGCN) for task prediction with sparse historical inputs.

\subsubsection{Multi-Scale Time Series Feature Fusion Block}

Considering that the intrinsic source of task flow generation in warehousing scenarios stems from the order requests of multiple merchants, we can directly and effectively extract the temporal dependencies of task flow by decoupling different order patterns and certain anomalous orders. Specifically, we employ a Fast Fourier Transform method to extract the dominant $k_{top}$ frequency components $\omega_i(i=1,2,...,k_{top})$, aiding the network in identifying periodic patterns\cite{timesnet}, such as fluctuations in seasonal delivery demand. Simultaneously, we utilize Discrete Wavelet Transformation to decompose the waveform coefficients at different frequency bands\cite{wavelet}, enabling the analysis of local characteristics of $\textbf{X}_{raw}$ at various temporal scales. This decomposition helps to identify patterns such as sudden increases in delivery demand. Subsequently, referencing \cite{pdformer}, we reconstruct the multi-scale time series to ${\widetilde{\textbf{X}}_{raw}}$, which is a two-dimensional tensor with each dimension of $R$ and then capture the sequential temporal dependencies using a 2D CNN layer.

To address the issue of sparse task flow distribution mentioned in~\ref{subsec: vs}, we introduce a special ``domain transfer'' training mechanism, which is embodied in the network's \textit{embedded mechanism} and\textit{ reverse mapping} operation. Following RBF networks, we firstly transform sparse spatial-temporal data into a more continuous latent representation using a Gaussian kernel-based embedding mechanism:
\vspace{-5pt}
\begin{equation}
\Tilde{X} = \sum_{j=1}^k \exp\left(-\frac{\|X - C_j\|^2}{2\sigma^2}\right) W_j^T + b
\vspace{-2pt}
\end{equation}Where $C_j$ is the temporal center of series, and $\sigma$ is an adaptive parameter. The weight matrices $W_j^T$ are updated during training to minimize the loss between the original data $X$ and the reconstructed data $\Tilde{X}$, enhancing the model's ability to capture and retain essential data features even from highly sparse inputs.
The reverse mapping employs a fuzzy reconstruction approach to approximate the original data from the embedded representation:
\vspace{-5pt}
\begin{equation}
{X} = \sum_{j=1}^k \pi_j(\Tilde X) \cdot \exp\left(-\frac{\|\Tilde{X} - \mu_j(\Tilde X)\|^2}{2 \sigma_j^2(\Tilde X)}\right) \cdot W_j^{'}(\Tilde X) + \Tilde b
\vspace{-2pt}
\end{equation}
$\pi_j(\Tilde X)$ are dynamically adjusted mixing coefficients, conditioned on the encoded data $\Tilde X$. The weights $W_j^{'}(\Tilde X)$ are tailored through the training process to accommodate the inherent variability and uncertainty of the embedded representations.

\subsubsection{3D Spatio-Temporal GCN Block}
\begin{figure}[htpb]
    \centering
    \makebox[\linewidth]{\includegraphics[width=0.95\linewidth]{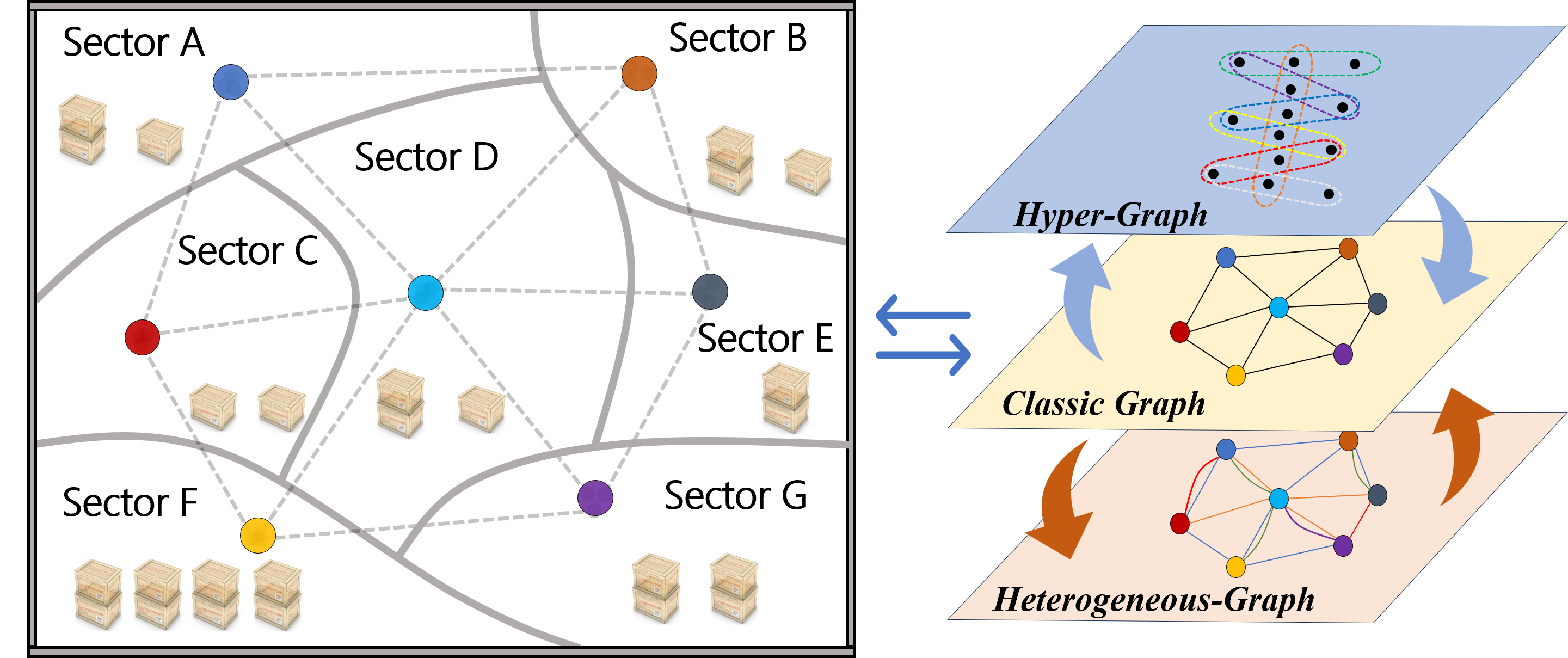}}
    \caption{The predicted tasks are distributed such that they appear in each sector of $G_H$, rather than being specific to each node within $G$(left). And their spatio-temporal distribution dependencies are extracted through iterative extraction and integration using three types of graph structures(right).}
    \vspace{-0.3cm}
    \label{fig:task distribution_1}
\end{figure}

The distribution of predicted tasks in $G_H$ is shown on the left of Fig.~\ref{fig:task distribution_1}. To address the challenges related to the complex adjacencies between sectors in the warehousing system, which are hard to characterize as discussed in Section~\ref{subsec: vs}, we propose a 3D Spatio-Temporal GCN Block that integrates spatio-temporal diffusion graph convolution \cite{dgcn}, hypergraph convolution \cite{hypergraph}, and heterogeneous graph convolution \cite{hgcn_1}. The relationships among them, as well as the connections with the tasks distribution are illustrated as Fig.~\ref{fig:task distribution_1}. 

The 3D GCN Block meticulously accounts for the real-world spatial-temporal dependencies inherent in task distribution. It specifically considers: \textit{i) the relationships among the static locations of cargoes within the warehouse, ii) the dynamic transportation of cargoes across the road network between sectors, and iii) the various types of interactions between sectors}. These dependencies can influence the emergence of future tasks in specific patterns across different sectors. Each factor is carefully analyzed to extract potential spatial information within a defined spatial dimension.

Before preforming each graph convolution operation, we employ a G-TCN \cite{gwn}, utilizing dilated convolution to further capture temporal dependencies within the task flow. Given the input $\mathcal{I}^{\mathbb{R} \times L \times F}$, G-TCN is formulated as:
\vspace{-2pt}
\begin{equation}
    G\text{-}TCN(\mathcal{I}) = \sigma_1(TCN_1(\mathcal{I})) \odot \sigma_2(TCN_2(\mathcal{I}))
\vspace{-2pt}
\end{equation}
where $G\text{-}TCN$($\mathcal{I}$) $\in \mathbb{R}^{R \times (L-b(t'-1)) \times F}$, dilated convolution is with a kernel $k_e \in \mathbb{R}^{t'}$ and dilation factor is $b$. $\sigma_1$ and $\sigma_2$ are activation function to make it nonlinear.


Following \cite{gwn}, we designe the Diffusion Graph Convolution Module to capture directed connections between sectors. To simulate the diffusion process of node information over $N$ steps, let $\Lambda_f = \textbf{A}/\text{\text{rows}}(\textbf{A})$ represent the forward transition matrix, and $\Lambda_b = \textbf{A}^T/\text{\text{rows}}(\textbf{A}^T)$ represent the backward transition matrix. Then D-GCN can be expressed as:
\vspace{-3pt}
\begin{equation}
    D\text{-}GCN(\mathcal{X}) = \sum_{i=0}^{N-1}\big(\Lambda_b^i \mathcal{X} \Theta_1^k + \Lambda_f^i \mathcal{X} \Theta_2^k + \Tilde{A}_{adp} \mathcal{X} \Theta_3^k \big)
\vspace{-5pt}
\end{equation} where $\Tilde{A}_{adp}$ is the adaptive adjacency matrix with learnable parameters, as also mentioned in the work. 

As for Double H-Graph Convolution Module, prior to entering its two sub-convolution module, the input graph structure undergoes dual transformation or heterogenization, respectively, to adapt to the specific convolutions. 

For the dual transformation in the diffusion hypergraph convolution module, following the approach in \cite{ddstgcn}, the incidence matrix $\textbf{H}$ between the graph $G_H$ and the hypergraph $\mathcal{G}_H$ encapsulates information from both structures. Therefore, this transformation can be accomplished through the following operations, $G_H \rightarrow \mathcal{G}_H : \text{ } \Tilde{\mathcal{X}_h} = (W\bigodot \textbf{H} \mathcal{X}_h) \bigoplus \mathcal{X}_{h,dist}$, and $\mathcal{G}_H \rightarrow G_H : \text{ } \mathcal{X}_h = (W'\bigodot \textbf{H})^T \mathcal{X}_h$, where $\bigoplus$ means the operartion of concentration, and $\mathcal{X}_{h,dist} \in \mathbb{R}^\gamma$ is the dist matrix of the high-level warehouse roadmap. Thus, from the update feature $\mathcal{X}_h \in \mathbb{R}^{\lambda \times \lambda}$, the convolution for feature extraction in the hypergraph $\mathcal{G}_H$ is given by:
\vspace{-4pt}
\begin{equation}
    DH\text{-}GCN(\mathcal{X}_h) = \sum_{i=0}^{N-1}(\textbf{D}_{v}^{-\frac{1}{2}} \textbf{H} \Psi_{adp}^i \textbf{D}_{e}^{-\frac{1}{2}} \textbf{H}^T \textbf{D}_{v}^{-\frac{1}{2}}) \mathcal{X}_h \Theta_h^i
    \vspace{-4pt}
\end{equation}
where $\textbf{D}_{e} \in \mathbb{R}^{\gamma \times \gamma}$ and $\textbf{D}_{v} \in \mathbb{R}^{\lambda \times \lambda}$ represent the diagonal matrices corresponding to the edge degrees and node degrees of $\mathcal{G}h$, respectively. $\Psi_{adp} \in \mathbb{R}^{\gamma \times \gamma}$ denotes a diagonal matrix that includes adaptive learning parameters, while $\Theta_h$ represents the parameter to be learned.
In Hetero-GCN module, the convolutional method is an extension of D-GCN tailored for heterogeneous graph $\mathcal{G}_{hetero}$. Heterogenization transforms the adjacency matrices based on the relationship tensor $\mathbb{F} \in \mathbb{R}^{\lambda \times \lambda}$ reflecting interactions within different sectors, leading to a redefined adjacency matrix $\textbf{A}_{HG} = \{\textbf{A}_{i,j} \mid i \in [1,m], j \in [1, n]\}$, where $m$ and $n$ represent the types of nodes and edges within the heterogeneous graph\cite{hgcn_1}. This matrix incorporates all heterogeneous information by representing different node and edge types. Thus when spatially dependent information is diffusely propagated in a graph, the effects of constraints on edge types are taken into account by colleagues. After this, we aggregate the nodes' multi-dimensional features as the high-dimensional feature need to be reduced for data concatenation. This operation can be effectively achieved using principal component analysis, which transforms the features into a normalized feature matrix while preserving essential information.

Based on the GCN architecture as described, the 3D GCN blocks are interconnected via residual connections, which are strategically designed to maintain the flow of information across layers and prevent the degradation of learned features. These residual connections allow the model to effectively capture and preserve critical spatial information to learn more complex patterns within the task flow, ultimately leading to more accurate and general predictions for future tasks.
\vspace{-2pt}
\subsection{Multi-Robot Hybrid Tasks Allocation}
We develope a heuristic algorithm $Hybrid\text{-}KM$ for hybrid tasks allocation based on the Hungarian algorithm \cite{ha}. We mainly introduce heuristic enhancements to the cost function, followed by dynamic bipartite graph matching. For a set of hybrid tasks $T_h$ and the set of robots $R$, when considering the matching cost between each idle robot $r_j$ and a specific task $T_i$, the heuristic cost matrix is given by:
\vspace{-4pt}
\begin{equation}
    C_{i,j} = \alpha \cdot u(T_i) + \beta \cdot d_{h}(r_j,T_i)+\sigma \cdot \eta_{S_i}
    \vspace{-4pt}
\end{equation}
in which $\alpha, \beta, and\text{ }\sigma \geq 0$ are normalization factors, and $\alpha=0$ if $T_i \in T^{real}$.
The $d_h$ refers to the the abstract distance cost from $r_j$ to $T_i$\cite{dynamic_allocation}. 
Variant $\eta_{S_i}$ represents the real-time task completing rate within the given sector $S_i$ containing task $T_i$. And the information relevant to $u(T_i)$ is derived from the uncertainty provided by predictive model, expressed as: $u(T_i) = \mu \cdot Con(T_i) - (1-\mu) \cdot EnScore(\textbf{X}_{raw})$, where $Con(\cdot)$ denotes the confidence and $EnScore(\cdot)$ indicates an entropy evaluation of the incremental historical tasks. 

Based on our $Hybrid\text{-}KM$ algorithm, the hybrid tasks are dynamically allocated in time. Specifically, whenever the predictor makes a new round of prediction, the updated tasks $T^u_h$ is matched with $R$ once. If there are enough tasks, idle robots may be pre-scheduled to act ahead, in addition to which, working robot $r^w$ may be re-assigned due to a smaller cost $C_{w,p}$ with a predicted task $T^p$. At each timestep, allocator tracks whether the predicted tasks have been officially published. The given robots will not be allocated new tasks untill the predicted horizon is reached. However, if we discover $T^f$ falsely predicted before next prediction, the robot will be re-assigned to other tasks in the next round. These measures lead to an increase in robot utilization. Furthermore, dynamic volatile of sector-level task completion rates drives robots to move towards sectors with lower scores. Consequently, the allocation results tend to achieve a more balanced distribution of robots.

Our main purpose of this work is to develop a general framework which exploits future tasks to explore the potentials of warehousing scheduling abilities. If necessary, any other prediction and pre-assignment methods can be deployed within our framework, and one can always introduce more complex model structures in each part of our framework to further improve the performance.



\vspace{-6pt}
\section{Implementation}

\subsection{Datasets of Task Flow}
Our experiments are conducted using three datasets, which are derived from real-world factories. These datasets, denoted as \texttt{TG}, \texttt{JA}, and \texttt{BB}, document the task distribution within $G_H$ at half-minute intervals, resulting in a representative spatio-temporal flow dataset. They encompass task flow for 2 months, each presenting a distinct warehousing scenario.

\subsection{Comparison Models and Parameter Settings}
We compare with several mainstream models: \textit{STGCN}\cite{stgcn} is a critical benchmark method for spatio-temporal prediction, while \textit{Graph-WaveNet}\cite{gwn} combines GCN and WaveNet to enhance dynamic spatio-temporal modeling capabilities. \textit{DDSTGCN}\cite{ddstgcn} and \textit{STSGCN}\cite{stsgcn} employ dual dynamic mechanisms and synchronous spatio-temporal graph convolution respectively, further improving the ability to capture spatio-temporal patterns. While \textit{D2STGNN}\cite{d2stgnn} decouples the diffusion signals and intrinsic signals within the spatio-temporal data to imporve performance.To account for the sparse attributes of task flow, we integrate our embedded mechanism with these backbones for fairness. 

All experiments are conducted on a computer with an Intel Core i5-13600K CPU, 32 GB of RAM, and GTX 4070 GPU. We use the Adam optimizer, with a learning rate of 5e-4, weight decay of 1e-4, and a dropout rate of 0.3. For our TDTGCN model, we configure four main blocks with a G-TCN dilation factor alternating between 1 and 2. The datasets are manually split into train, test, and validation sets in a 7:2:1 ratio. Each model inputs data from the past 12 frames to predict task flow for future intervals of 3, 5, 10, and 15.

\subsection{Evaluation Metrics}
We use the following key metrics to evaluate the performance of our pre-scheduling enhanced framework in warehousing system:

I) Empty Running Rate (ERR)
\begin{equation}
ERR = \frac{\sum_{i=1}^{N_r}\sum_{j=1}^{K}t_{p,i}^j}{\sum_{i=1}^{N_r}{\sum_{j=1}^{K}(t_{p,i}^j + t_{d,i}^j)}}
\end{equation}
where $K$ is the total number of time steps, and $N_r$ is the number of working robots during $K$ time steps. $t_{p,j}=1$ if $r_i$ is heading for a task or just idle, and $t_{d,i}=0$ at the same time, vice versa.

II) Mean Pickup Time (MPT)
\begin{equation}
MPT = \frac{1}{N_r}\sum_{i=1}^{N_r}\sum_{j=1}^{K}t_{p, i}^j    
\end{equation}

III) Misguided Trip Ratio (MTR)
\begin{equation}
MTR = \sum_{i=1}^{N_r}\frac{D_{misled}^{i}}{D_{total}^i}
\end{equation}
where $D_{misled}^i$ represents the additional distance $r_i$ is misled to travel. This metric reflects the proportion of additional paths due to accidental mispredictions.

The performance of prediction is assessed using common metrics: Mean Absolute Error (MAE), Root Mean Square Error (RMSE), and Weighted Mean Absolute Percentage Error (WMAPE), as detailed in~\cite{han_1}.

\subsection{Large-Scale Validation Experiment Settings}
To validate the scalability, we execute large-scale experiments across three real maps of different sizes, denoted as $\mathcal{M}_L, \mathcal{M}_M, \mathcal{M}_S$. They are also derived from actual projects, and the settings are detailed in Table~\ref{tab:sim_settings}. We adjust the prediction horizons to 5, 10, and 15 to assess the performance and the extra cost (present by MTR) with each horizon respectively. To ensure the credibility, we generate 5 different scenarios for each map and execute for 5 times.
\begin{table}[htbp]
  \centering
  \caption{Instructions of Large-Scale Scenarios}
  \setlength{\tabcolsep}{6pt}
  \renewcommand{\arraystretch}{1.2}
  \label{tab:sim_settings}
    \begin{tabular}{c|cccc}
    \hline
    \hline
    Maps   & nodes & edges & sectors & tasks \\
    \hline
    \textit{\textbf{large map}} $\mathcal{M}_L$   & 10155 & 34616 & 29 & 12793 \\
    \textit{\textbf{middle map}} $\mathcal{M}_M$    & 5862  & 23553 & 19 & 6398 \\
    \textit{\textbf{small map}} $\mathcal{M}_S$   & 2745  & 9129  & 6 & 3601 \\
    \hline
    \hline
    \end{tabular}%
  \label{tab:intructions}%
\end{table}%

\begin{table*}[htbp]
  \centering
  \caption{Comparison of Prediction Results Based on Real Task Flow Datasets}
  \label{pred_result}
  \resizebox{1\textwidth}{!}{
    \begin{tabular}{c|c|ccc|ccc|ccc|ccc}
    \hline
    \hline
    \multirow{2}[1]{*}{\textbf{DataSets}} & \multirow{2}[1]{*}{\textbf{Models}} & \multicolumn{3}{c|}{\textbf{Horizon 3}} & \multicolumn{3}{c|}{\textbf{Horizon 5}} & \multicolumn{3}{c|}{\textbf{Horizon 10}} & \multicolumn{3}{c}{\textbf{Horizon 15}} \\
\cline{3-14}          &       & MAE$\downarrow$   & RMSE$\downarrow$  & WMAPE$\downarrow$ & MAE   & RMSE  & WMAPE & MAE   & RMSE  & WMAPE & MAE   & RMSE  & WMAPE \\
    \hline
    \multirow{6}[2]{*}{\texttt{TG}} & STGCN & 0.4313  & 0.6452  & 0.2603  & 0.5764  & 1.1144  & 0.3497  & 0.8388  & 1.5438  & 0.4610  & 1.1352  & 1.8182  & 0.6126  \\
          & Graph WaveNet & 0.4115  & 0.6367  & 0.2503  & 0.6007  & 1.1852  & 0.3614  & 0.8389  & 1.6422  & 0.4819  & 1.0610  & 1.7671  & 0.5827  \\
          & DDSTGCN & 0.4065  & 0.6015  & 0.2256  & 0.5695  & 1.0744  & 0.3163  & 0.8234  & 1.4982  & 0.4577  & 1.0191  & 1.7714  & 0.5667  \\
          & STSGCN & \underline{0.3642}  & \underline{0.5971}  & \underline{0.2149}  & 0.5218  & \underline{0.9873}  & 0.3097  & \underline{0.7769}  & \underline{1.3764} & \underline{0.4256} & \underline{0.9694}  & \underline{1.6691}  & \textbf{0.5616} \\
          & D2STGNN & 0.3957  & 0.6594  & 0.2319  & \underline{0.5170}  & 1.0299  & \underline{0.2972}  & 0.7869  & 1.4746  & 0.4263  & 0.9974  & 1.7113  & 0.5741  \\
          & \textit{\textbf{ours}} & \textbf{0.3535} & \textbf{0.5714} & \textbf{0.2036} & \textbf{0.4829} & \textbf{0.8823} & \textbf{0.2742} & \textbf{0.7550} & \textbf{1.3701}  & \textbf{0.4258}  & \textbf{0.9441} & \textbf{1.6876} & \underline{0.5644}  \\
    \hline
    \multirow{6}[2]{*}{\texttt{JA}} & STGCN & 0.4050  & 0.6229  & 0.2328  & 0.6184  & 0.8519  & 0.3229  & 0.7035  & 1.3966  & 0.4695  & 0.9848  & 1.7215  & 0.6361  \\
          & Graph WaveNet & 0.4282  & 0.6243  & 0.2461  & 0.6801  & 0.9793  & 0.3665  & 0.8646  & 1.4241  & 0.4883  & \underline{0.8373}  & 1.5523  & 0.5372  \\
          & DDSTGCN & 0.3911  & 0.5968  & 0.2336  & 0.5263  & 0.8284  & 0.2748  & 0.6936  & 1.3682  & 0.4412  & 0.8762  & \underline{1.5036} & 0.5022  \\
          & STSGCN & \underline{0.3612}  & \underline{0.5439}  & \underline{0.2158}  & \underline{0.4493}  & \textbf{0.7733} & \underline{0.2687}  & \underline{0.6775}  & \underline{1.2754}  & \underline{0.4059}  & 0.8254  & 1.5112  & \underline{0.4951}  \\
          & D2STGNN & 0.3772  & 0.5689  & 0.2207  & 0.4548  & 0.8314  & 0.2866  & 0.6972  & 1.3763  & 0.4360  & 1.0340  & 1.7531  & 0.6194  \\
          & \textit{\textbf{ours}} & \textbf{0.3518} & \textbf{0.5281} & \textbf{0.2101} & \textbf{0.4251} & \underline{0.7625}  & \textbf{0.2601} & \textbf{0.6659} & \textbf{1.2603} & \textbf{0.3989} & \textbf{0.8141} & \textbf{1.4939}  & \textbf{0.4882} \\
    \hline
    \multirow{6}[2]{*}{\texttt{BB}} & STGCN & 0.4893  & 0.8078  & 0.2461  & 0.5957  & \underline{0.9992}  & 0.3864  & 0.7324  & 1.2480  & 0.5226  & 1.2761  & 1.4822  & 0.6394  \\
          & Graph WaveNet & 0.4913  & 0.7946  & 0.2134  & 0.6641  & 1.0946  & 0.3941  & 0.7625  & 1.5085  & 0.5688  & 0.9849  & 1.5334  & 0.5433  \\
          & DDSTGCN & 0.4408  & 0.8134  & 0.1944  & 0.5764  & 1.0271  & 0.3646  & \underline{0.7235}  & \textbf{1.1275} & \underline{0.5156}  & 0.9706  & \underline{1.4581}  & 0.5371  \\
          & STSGCN & \underline{0.3934}  & \underline{0.7693}  & \underline{0.1735}  & 0.5607  & 1.1956  & \underline{0.3473}  & 0.7472  & 1.3746  & 0.5345  & \underline{0.9579}  & 1.5067  & \underline{0.5191}  \\
          & D2STGNN & 0.4187  & 0.7787  & 0.1764  & \underline{0.5439}  & 1.1421  & 0.3648  & 0.7822  & 1.2013  & 0.5437  & 1.0394  & 1.5174  & 0.5261  \\
          & \textit{\textbf{ours}} & \textbf{0.3656} & \textbf{0.7238} & \textbf{0.1524} & \textbf{0.5167} & \textbf{0.9453} & \textbf{0.3246} & \textbf{0.7113} & \underline{1.1801}  & \textbf{0.4719} & \textbf{0.9462} & \textbf{1.4475} & \textbf{0.5046} \\
    \hline
    \hline
    \end{tabular}%
    }
    \begin{tablenotes}
        \footnotesize
        \item[\dag] \textbf{Bolding} indicates best performance, \underline{underlining} indicates second best.
    \end{tablenotes}
  \label{tab:addlabel}%
\end{table*}%

\vspace{-10pt}

\begin{table*}[htpb]
    \centering
    \caption{Results of Large-Scale Validation Experiments}
    \label{sim_result}
    \begin{tabular}{m{1cm}|c|c|ccc|c|ccc|c|ccc}
    \hline
    \hline
    \multirow{3}{*}[-0.5em]{\makecell{\textbf{Scens}\\ \& \\\textbf{Settings}}}
    ~  & \textit{Scens}
     & \multicolumn{4}{c|}{$\mathcal{M}_L$ with \textit{\textbf{50} robots and \textbf{12793} tasks}}
      & \multicolumn{4}{c|}{$\mathcal{M}_M$ with \textit{\textbf{20} robots and \textbf{3601} tasks}}
       & \multicolumn{4}{c}{$\mathcal{M}_S$ with \textit{\textbf{30} robots and \textbf{6398} tasks}}  \\
    \cline{2-14}
    
     ~ & \multirow{2}{*}{\textit{Horizon/30s}}  & {classic} & \multicolumn{3}{c|}{pred-enhanced} & {classic} & \multicolumn{3}{c|}{pred-enhanced} & {classic} & \multicolumn{3}{c}{pred-enhanced} \\
    \cline{3-6} \cline{7-10} \cline{11-14}
    ~ &~ \rule{0pt}{2.0ex}& - & 5 & 10 & 15 & - & 5 & 10 & 15 & - & 5 & 10 & 15 \\
    \hline
    \multirow{3}{*}{\centering \textbf{Metrics}}
    \rule{0pt}{2.3ex}& \textbf{ERR}(\%)$\downarrow$
        & \textit{36.61} & 28.51 & 23.67 &20.60 & \textit{33.84} & 25.44 & 21.37 & 18.05 & \textit{32.04} & 20.94 & 17.06 & 14.13 \\
        \cline{2-14}
        \rule{0pt}{2.32ex} &   \textbf{MPT}(mins)$\downarrow$
     & \textit{5.949} & 4.637 & 3.837 &3.344  & \textit{3.806} & 2.860 & 2.397 & 2.025 & \textit{5.082} & 3.318 & 2.704 & 2.243 \\
     \cline{2-14}
     \rule{0pt}{2.5ex} &  \textbf{MTR}(\%)$\downarrow$
    \rule{0pt}{1.5ex} & \textit{0} & 0.28 & 0.88 & 1.06 & \textit{0} & 0.14 & 0.55 & 0.63 & \textit{0} &  0.17 & 0.41 & 1.25  \\
    \hline
    \hline
    \end{tabular}
    \begin{tablenotes}
        \footnotesize
        \item[\dag]- means the value is \textbf{None}, as it pertains to traditional scheduling.
    \end{tablenotes}
        \vspace{-13pt}
\end{table*}

\section{Real-World Applications}

\subsection{Evaluation of Task Flow Prediction}
The result of  prediction experiments, shown in Table \ref{pred_result}, indicate that our method consistently outperforms others. For short-term predictions (horizons of 3 and 5), our network achieves SOTA performance, with improvements of up to 7.07\%, 10.64\%, and 9.16\% in MAE, RMSE, and WMAPE, respectively. In mid-term predictions (horizon of 10), our method shows gains of up to 2.8\%, 5.3\%, and 8.47\%. Although the improvement margins narrow, our approach still generally outperforms others. For long-term predictions (horizon of 15), ours are still ahead in most cases, although there is an effect of error accumulation.

\begin{figure}[htpb]
    \centering
    \makebox[\linewidth]{\includegraphics[width=0.8\linewidth]{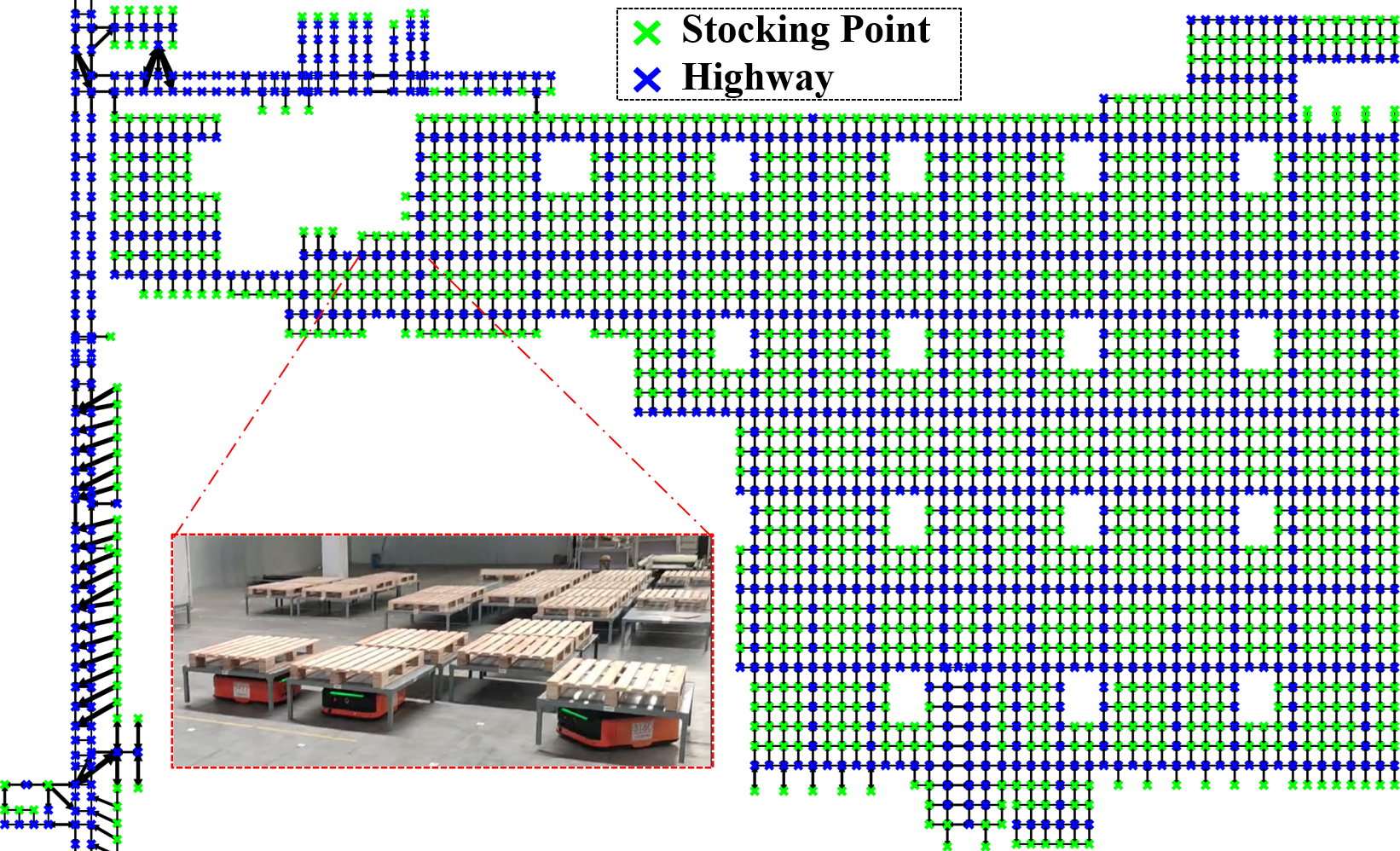}}
    \caption{The layout of real-world factory for validation experiments}
    \label{fig: map_tg}
    \vspace{-16pt}
\end{figure}

\subsection{Result of Real-World Lifelong Validation}
By deploying task flow prediction, we apply our approach in industrial project, from which \texttt{TG} is derived. The scenario is illustrated in Fig.~\ref{fig: map_tg}. We divide the topology of factory into two layers: the high-level map comprises highways and inventory sectors (indicated by the green areas), while the low-level map is the actual road network topology of warehouse. We conduct experiments for months with 3 different allocation strategies, and the results are shown in~\ref{tab:lifelong}. It's shown that our approaches significantly increase the efficiency of MRS, only losing 0.98\% of the path optimality. 

\section{Large-Scale Validation Results}
Table~\ref{sim_result} presents the results of large-scale validation experiment, demonstrating obvious improvements under large-scale robotic warehousing system,  while maintaining path optimality within about 1\% loss.

\begin{table}[htbp]
  \centering
  \caption{Life-Long Validation Results on Map \texttt{TG}}
    \begin{tabular}{c|ccc}
    \hline
    \hline
    Methods of Allocation   & \multicolumn{1}{l}{ERR(\%)$\downarrow$} & \multicolumn{1}{l}{MPT(mins)$\downarrow$} & \multicolumn{1}{l}{MTR(\%)$\downarrow$} \\
    \hline
    \textit{Greedy} & 57.4  & 5.96  & 0 \\
    \textit{Classic}\cite{ha}  & 46.1  & 3.98  & 0 \\
    \textbf{Ours}  & \textbf{31.9}  & \textbf{2.71}  & \textbf{0.98} \\
    \hline
    \hline
    \end{tabular}%
  \label{tab:lifelong}%
  \vspace{-16pt}
\end{table}%

Our approach significantly reduces ERR, with reductions exceeding 50\%. As the prediction horizon extends, ERR continues to decrease, albeit at a slower rate. For instance, in the scenario $\mathcal{M}_S$, ERR drops from 32.04\% to 14.13\% at 15 horizon, reflecting a 55.91\% improvement.
Additionally, MPT shows substantial reductions across all three maps. In the largest map, $\mathcal{M}_L$, with 50 robots and 12,793 tasks, it decreases by 22.1\%, 35.5\%, and 43.8\% across increasing prediction horizons. Although the reduction rate diminishes, the improvements remain highly satisfactory. In the other two maps, even greater MPT reductions are observed; for example, in $\mathcal{M}_S$, it decreases by 55.8\% at 15 horizon.

It is worth noting that MTR slightly increases as horizon extends, but still stays at or below about 1\% while achieving 50-60\% improvements in ERR and MPT. Notably, in the smallest and simplest map $\mathcal{M}_M$, the MTR values are 0.14\%, 0.55\%, and 0.63\% at different horizons. 


\vspace{-3pt}
\section{Conclusion}
We propose TDTGCN to predict task flow, upon which we develope a pre-scheduling enhanced framework for multi-robot warehousing system. It allocates tasks more reasonably by uncovering implicit information within the MRS. Real-world applications with hundreds of robots validate our feasibility.
We plan to integrate dynamic sector partitioning with task flow prediction in the future.


\bibliographystyle{ieeetr} 
\bibliography{ref}

\end{document}